\documentclass[letterpaper]{article} 
\usepackage{aaai25}  
\usepackage{times}  
\usepackage{helvet}  
\usepackage{courier}  
\usepackage[hyphens]{url}  
\usepackage{graphicx} 
\usepackage{natbib}  
\usepackage{caption} 
\usepackage{algorithm}
\usepackage{algorithmic}
\usepackage{amsmath}
\usepackage{multirow}
\usepackage{subcaption}

\usepackage{newfloat}
\usepackage{listings}
\DeclareCaptionStyle{ruled}{labelfont=normalfont,labelsep=colon,strut=off} 
\floatstyle{ruled}
\newfloat{listing}{tb}{lst}{}
\floatname{listing}{Listing}
\title{TSVC:\,Tripartite Learning with Semantic Variation Consistency for Robust Image-Text Retrieval}
\author {
    Shuai Lyu\textsuperscript{\rm 1},
    Zijing Tian\textsuperscript{\rm 2},
    Zhonghong Ou\textsuperscript{\rm 3}\thanks{Corresponding author.},
    Yifan Zhu\textsuperscript{\rm 1},
    Xiao Zhang\textsuperscript{\rm 1},
    Qiankun Ha\textsuperscript{\rm 1},
    Haoran Luo\textsuperscript{\rm 1},
    Meina Song\textsuperscript{\rm 1}
}
\affiliations {
    \textsuperscript{\rm 1}School of Computer Science, Beijing University of Posts and Telecommunications, China\\
    \textsuperscript{\rm 2}School of Science, Beijing University of Posts and Telecommunications, China\\
    \textsuperscript{\rm 3}State Key Laboratory of Networking and Switching Technology, Beijing University of Posts and Telecommunications, China\\
    \{Lxb\_savior, tianzj, zhonghong.ou, yifan\_zhu, xiao20010420\}@bupt.edu.cn
}

\usepackage{bibentry}

\begin{document}

\maketitle

\begin{abstract}
Cross-modal retrieval maps data under different modality via semantic relevance. 
Existing approaches implicitly assume that data pairs are well-aligned and ignore the widely existing annotation noise, i.e., noisy correspondence\,(NC). Consequently, it inevitably causes performance degradation. 
Despite attempts that employ the co-teaching paradigm with identical architectures to provide distinct data perspectives, the differences between these architectures are primarily stemmed from random initialization. Thus, the model becomes increasingly homogeneous along with the training process.
Consequently, the additional information brought by this paradigm is severely limited. 
In order to resolve this problem, we introduce a Tripartite learning with Semantic Variation Consistency\,(TSVC) for robust image-text retrieval. We design a tripartite cooperative learning mechanism comprising a Coordinator, a Master, and an Assistant model. The Coordinator distributes data, and the Assistant model supports the Master model's noisy label prediction with diverse data. Moreover, we introduce a soft label estimation method based on mutual information variation, which quantifies the noise in new samples and assigns corresponding soft labels. We also present a new loss function to enhance robustness and optimize training effectiveness. Extensive experiments on three widely used datasets demonstrate that, even at increasing noise ratios, TSVC exhibits significant advantages in retrieval accuracy and maintains stable training performance.

\end{abstract}

%

\section{Introduction}

Cross-modal retrieval~\citep{DBLP:conf/cvpr/00010BT0GZ18,DBLP:journals/tetci/LiTLF19} aims to accurately associate and align data from different modalities, e.g., images and texts. As a key technology in the field of multi-modality, it has been widely applied in both industry and academia. In classification tasks, noise labels \cite{yan2023adaptive,iscen2022learning,yan2022noise} generally refer to labeling errors. In cross-modal matching tasks, however, noise labels pertain to alignment errors in paired data, also known as noise correspondence. Consequently, a large fraction of existing noise-robust learning methods~\cite{DBLP:conf/icml/ZhongWZY024} designed for classification cannot be directly applied for cross-modal matching tasks.

\begin{figure}[t]
    \centering
    \includegraphics[width=\linewidth]{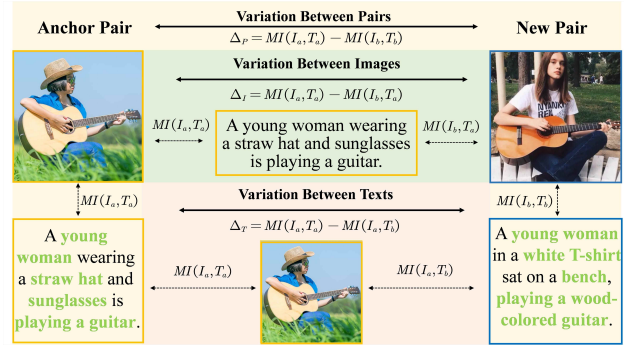}
    \caption{Illustration of semantic variance consistency guided sample filtering. We use Mutual Information\,(MI) to calculate the level of noise between anchor pair and new pair. We primarily consider the MI variation among sample pairs $\Delta_P$, images $\Delta_I$, and texts $\Delta_T$. The smaller the variation is, the cleaner we consider the pair to be.}
    \label{fig:image1}
\end{figure}

To address noisy correspondences in cross-modal matching, existing studies mainly focus on two aspects: noisy sample label estimation~\cite{ma2024cross,huang2024cross} and loss function adjustment~\cite{hu2021learning,shibreaking}. Noisy sample label estimation primarily involves re-estimating and adjusting the labels of noisy samples by exploring the latent relationships within the dataset. The loss function adjustment aims to enhance model robustness by designing new loss functions. Recent studies have increasingly focused on the estimation of noisy sample labels.

Accurately estimating soft corresponding labels for noisy data has always been a big challenge for noise-robust cross-modal matching. Previous studies~\cite{yang2024robust, zhao2024mitigating} have made efforts by employing the rate of similarity changes to quantify noise content. Yang et al.~\cite{yang2023bicro} leverage the inherent similarity between the two modalities to assign pseudo labels. Nevertheless, they fail to consider other crucial data relationships and features, thus limiting their ability to identify complex noise.
On the other hand, some studies~\cite{zheng2021meta,zhang2024negative} mainly rely on the memory effect of Deep Neural Networks\,(DNN)~\cite{zhang2021understanding}, which tends to prioritize learning simple patterns rather than noisy samples. They divide the training data into a clean set and a noisy set, and use the Co-Teaching training scheme~\cite{han2018co,yan2023adaptive} to train them with different mechanisms.

Nevertheless, in the Co-Teaching paradigm, the differences between two networks with the same architecture primarily arise from random initialization. During the training process, both sides provide what they deem as important data to each other for training, resulting in limited additional information gain.
Moreover, the predominant approaches \cite{yang2023bicro,qin2024cross,yang2024robust} for noise-robust tasks typically used minimizing triplet loss with a soft margin, which fails to consider the differences and distribution characteristics between clean and noisy samples, causing suboptimal performance in distinguishing between them.

In order to resolve the problems mentioned above, we propose a \textbf{T}ripartite learning with \textbf{S}emantic \textbf{V}ariation \textbf{C}onsistency (\textbf{TSVC}) scheme for robust image-text retrieval, which mainly consists of three components. 

First, we propose a \textbf{S}emantic \textbf{I}nformation \textbf{V}ariation \textbf{C}onsistency (\textbf{SIVC}) method for label estimation method, based on semantic variation of Mutual Information (MI) between new pairs and clean pairs. As shown in Fig.~\ref{fig:image1}, we take into consideration three parts, i.e., pairs, images, and texts. The smaller the change, the closer the MI of new pairs is to that of clean pairs. It indicates that the new samples are cleaner, leveraging broader data relationships to better identify and quantify noise for label estimation. 
Second, to avoid model homogenization, we propose a novel \textbf{Tri}partite Cooperative \textbf{Learning} Mechanism (\textbf{Tri-learning}) that deviates from the conventional Co-Teaching paradigm. Specifically, Tri-Learning consists of three models: a Coordinator, a Master, and an Assistant. The Coordinator partitions data into clean and noisy sets. The Assistant selects clean samples from the noisy set to enhance the Master model. The Master trains on diverse data while preserving the ability to extract clean samples.  The Coordinator adjusts partitioning for the next round based on feedback from the Assistant.
Third, we train the segmented dataset using the newly introduced \textbf{D}istribution-\textbf{A}daptive \textbf{S}oft \textbf{M}argin (\textbf{DASM}) loss function, which takes into account the dynamic margin affected by rectified soft label and sample distribution deviation. Our main contributions are summarized as follows:
\begin{itemize}
    \item We introduce a novel training paradigm Tri-learning to establish a collaborative relationship among three models. It mitigates the improvement limitation in traditional co-training paradigms, which is caused by the homogenization of models. 
    
    \item We propose a soft label estimation method namely SIVC based on semantic Mutual Information variation, and present a loss function called DASM that corrects margins and distribution deviations, which enhances noise detection accuracy and robustness significantly.
   
    \item We conduct extensive experiments on three cross-modal noise datasets, including synthetic and real-world noises. Experimental results show that TSVC significantly outperforms state-of-the-art methods.
\end{itemize}

\begin{figure*}[t]
    \centering
    \includegraphics[width=0.9\linewidth]{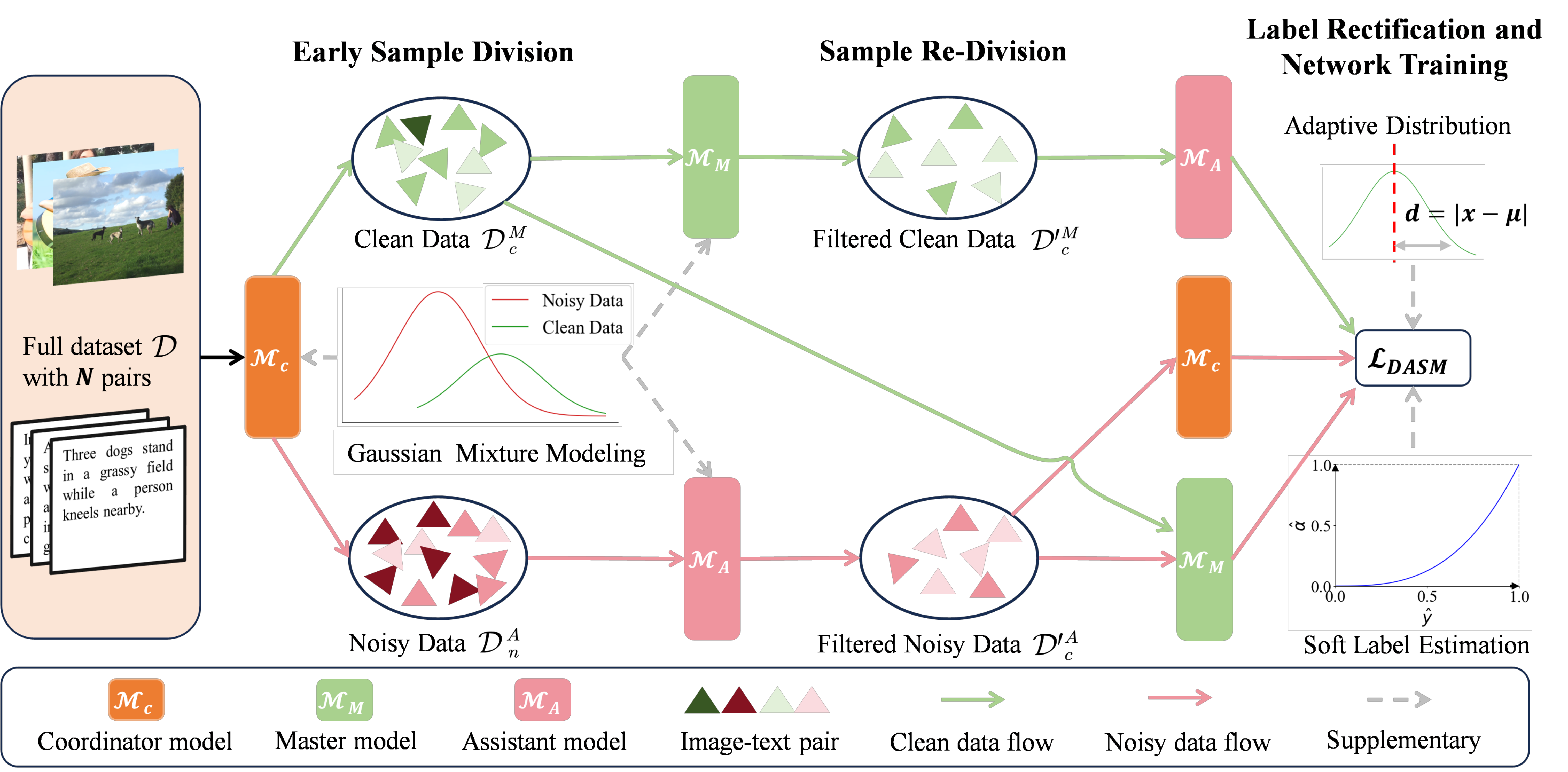}
    \caption{Illustration of Tripartite Cooperative Learning process.  We divide the full data into two flows. We first utilize the noisy data and filtered clean data to train the Assistant model. We then employ the filtered noisy data to train the Coordinator and the Master model. Note that the cleaner samples are lighter in color. The input to the three models consists of image-text sample pairs. During the division stage, the output is the DASM loss used by GMM for classification. In the inference stage, it is the predicted probability. }
    \label{fig:process}
\end{figure*}

\section{Related Work}
\subsection{Cross-modal Matching}
Cross-modal matching~\cite{yang2022dual,deng2019two,wang2020cross} is an fundamental problem in the fields of information retrieval~\cite{DBLP:conf/emnlp/ZhaoLZHCHZ24} and multi-modal analysis. It aims to retrieve one modality based on another modality as a query. 

Image-text retrieval, as the most common task, can be roughly divided into two categories, i.e., coarse-grained alignment and fine-grained alignment. Coarse-grained alignment~\cite{chen2021learning,faghri2017vse++,diao2021similarity} usually employs two separate networks to project the entire image and text into a unified embedding space. It then calculates the overall correlation using cosine similarity. Fine-grained alignment methods~\cite{lee2018stacked,chen2020uniter,zhang2022negative} often combine cross-modal interaction  with local segment alignment. Afterwards, they accumulate the scores of local regions to acquire the overall similarity~\cite{wu2019learning}. It is worth noting that the superior performance of these methods relies on a large amount of accurately annotated training data, without considering the issue of noise correspondence.

\subsection{Noisy Correspondence Learning}

Noise correspondence, different from noise label learning, usually refers to mismatched sample pairs in multi-modal datasets. Some studies~\cite{huang2021learning,han2023noisy,ma2024cross} leverage the memory effect of DNN to construct error correction networks to enhance the data purification process or identify mismatching samples. Other works~\cite{qin2022deep,yang2023bicro,han2024learning,li2024nac} improve the objective function with dynamic loss or bi-directional cross-modal similarity to identify the noise contained in the samples.

The methods mentioned above assume that the two networks can provide distinct data perspectives. Nevertheless, the additional information improvement remains limited, because the differences between the two networks with the same architecture primarily stem from random initialization~\cite{qin2024cross}. In contrast, we propose a Tripartite Cooperative Learning Method to augment network diversity, and utilize variations in Mutual Information across samples and modalities to estimate soft correspondence labels, which enhances the model's robustness against noise significantly.

\section{The Proposed Method TSVC}
\subsection{Problem Formulation}

Given the dataset $\mathcal{D} = \left( I_i, T_i, y_i \right)_{i=1}^N$ consisting of $N$ samples for training. Each of them includes one pair of image and text, as well as a binary label $y_i$ indicating the relevance of the pair as positive ($y_i=1$) or negative ($y_i=0$). In cross-modal matching, the primary objective is to maximize the similarity between positive samples (matched pairs) while minimizing the similarity between negative samples (unmatched pairs). 

Considering that multi-modal datasets are frequently annotated or acquired from Internet using cost-effective methods, it is inevitable to encounter noisy data in cross-modal matching tasks, which is known as the noisy correspondence problem. Specifically, it indicates that $(I_i, T_i)$ is a mismatched pair with corresponding label $y_i = 1$, which makes the model overfit to the noisy dataset and pull the negative samples closer, causing misleading results. To address this problem, we propose a framework called \textbf{Tripartite Learning with Semantic Variation Consistency\,(TSVC)} to enhance its robustness to noise. The details are outlined below.

\subsection{Semantic Information Variation Consistency} \label{sub:sivc}

\textbf{Mutual Information.} Existing studies~\cite{huang2024dynamic,wang2024balanced} suggest that clean samples have a stronger correlation between images and texts. Consequently, Semantic Information Variation Consistency\,(SIVC) leverages Mutual Information\,(MI)~\cite{shannon1948mathematical} to estimate soft correspondence labels. MI quantifies the shared information between signals, providing a criterion to measure the dependency between image-text pairs and determine sample noise proportion. For a pair of discrete sample pairs $(X, Y)$, MI is defined as:
\begin{equation}
    MI(X;Y) = \sum_{x \in X} \sum_{y \in Y} p(x, y) \log\left(\frac{p(x, y)}{p(x)p(y)}\right),
    \label{MI}
\end{equation}

where $p(x,y)$ indicates the joint probability distribution, $p(x)$ and $p(y)$ represent the marginal probability distributions.

We assume that $x$ and $y$ are vectors with $d$ dimension, we use the 1D histogram to approximate the marginal distributions. First, we divide the feature field into evenly spaced intervals, then count the frequency of each feature value $x_i$ falling within different intervals $[x_j,x_{j+1})$ to obtain $p(x)$:
\begin{equation}
    p(x)_ j = \frac{1}{d}\sum_{i=1}^{d} \delta(x_i\in [x_j,x_{j+1})),
\end{equation}


where $\delta(\cdot)$ is an indicator function. Same as $p(y)$.
Then, we use the 2D histogram to approximate the joint probability distribution $p(x,y)$. We divide the feature fields into a 2D grid, then we count the frequency of each feature pair $(x_i,y_i)$ falling within different cells to obtain $p(x,y)$:

\begin{equation}
    p(x,y)_ {j,k}=\frac{1}{d}\sum_{i=1}^d \delta(x_i \in[x_j,x_{j+1})) \times \delta (y_i \in [y_j,y_{j+1})).
\end{equation}




\noindent
\textbf{Semantic Consistency Calculation.} In practice, slight disparities exist in the similarity of clean subsets, and their labels may not exactly equal 1. When estimating soft correspondence labels, we prioritize variations in semantic consistency. Clean image-text pairs generally show balanced contributions from both modalities, avoiding excessive dominance by either.


First, we select a pair of samples $(I_a, T_a)$ as anchor points in each batch, similar to previous studies, by choosing the image-text pair with the minimum loss within the batch. These anchor points are considered representative of clean samples. Subsequently, for the given sample $(I_b, T_b)$, we perform calculations from three aspects to evaluate its consistency and noise characteristics.
\begin{itemize}
    \item Change rate of MI between image-text pairs:

\begin{equation}
   R_P=\frac{|MI\left( I_a,T_a \right) -MI\left( I_b,T_b \right)|}{MI\left( I_a,T_a \right)}.
\end{equation}

    \item Change rate of MI between texts:

\begin{equation}
   R_T=\frac{|MI\left( I_a,T_a \right) -MI\left( I_a,T_b \right)|}{MI\left( I_a,T_a \right)}.
\end{equation}

    \item Change rate of MI between images:

\begin{equation}
   R_I=\frac{|MI\left( I_a,T_a \right) -MI\left( I_b,T_a \right)|}{MI\left( I_a,T_a \right)}.
\end{equation}
\end{itemize}

\noindent
\textbf{Soft Label Estimation.} By calculating soft label estimation, we derive a standardized measure of the variations in consistency between given pairs and anchor pairs. Specifically, if the change rate between image-text pairs ($R_P$) is small, and the change rate between texts ($R_T$) is nearly equal to that between images ($R_I$), it indicates that texts and images contribute equally to the semantic information. In such cases, we infer that the new pairs are likely to be clean.

Eventually, we combine the inverse proportional function with the MI change rate to determine the soft correspondence label $y^* \in (0,1]$ for the given image-text pair:

\begin{equation}
   y^*=\frac{1}{1+(R_P+|R_T-R_I|)}.
\end{equation}

\subsection{Tripartite Cooperative Learning Mechanism}
We propose a framework named Tri-learning to overcome the limitations of traditional Co-training. The framework consists of three models: $\mathcal{M}_C$ (Coordinator)  partitions the original dataset into a clean set and a noisy set, $\mathcal{M}_A$ (Assistant) selects low-loss samples and combines them with the clean samples to train the $\mathcal{M}_M$ (Master). $\mathcal{M}_M$ further refines $\mathcal{M}_A$, while $\mathcal{M}_C$ iteratively adjusts the partitioning scheme for the next round based on the clean samples, which are fed back by the Assistant model from noisy samples. The detailed training process is illustrated in Fig.~\ref{fig:process} and is described as follows:

\textbf{Step 1: Early Sample Division.} For $\mathcal{M}_C$ , we adopt the same network structure as $\mathcal{M}_M$ and $\mathcal{M}_A$, with only different initialization parameters. With the assistance of GMM, $\mathcal{M}_C$ divides the original dataset into clean samples $\mathcal{D}_c^M$ with losses less than threshold $\delta$, and noisy samples $\mathcal{D}_n^A$ with losses larger than $\delta$ .

\textbf{Step 2: Sample Re-Division.} 
Through step 1, we obtain two separate data streams, one clean and the other noisy. For the clean data stream $\mathcal{D}_c^M$, we further mine the samples $\mathcal{D'}_{c}^{M}$ with losses less than  $\delta$  by the Master model and GMM. It ensures that the training samples of $\mathcal{M}_A$ are as clean as possible. For the noisy data stream $\mathcal{D}_n^A$, we use the Assistant model and GMM to select relatively clean data $\mathcal{D'}_c^A$  with losses less than  $\delta$, to provide diverse data for training  $\mathcal{M}_M$ and $\mathcal{M}_C$.

\textbf{Step 3: Label Rectification and Network Training.} 

We first apply the \textbf{SIVC} to rectify noisy labels within each training batch. Then, we train $\mathcal{M}_A$ using the cleaned dataset $\mathcal{D'}_c^M$ from step 2, equipping $\mathcal{M}_A$ with the ability to extract clean samples from the noisy dataset $\mathcal{D}_n^A$. Similarly, we optimize $\mathcal{M}_M$ using a combination of $\mathcal{D'}_c^A$ and $\mathcal{D}_c^M$, enabling it to gain valuable insights from diverse samples while maintaining robustness and generalization. 
To prevent error propagation during data filtering, $\mathcal{M}_C$ is iteratively optimized using $\mathcal{D'}_c^A$, as $\mathcal{D}_n^A$ may contain clean samples. This allows $\mathcal{M}_C$ to detect potential errors and adjust the training and predictions of $\mathcal{M}_A$ and $\mathcal{M}_M$.

We repeat Step 1, 2, and 3 for a specific number of iterations, then use the $\mathcal{M}_{A}$ and $\mathcal{M}_{M}$ models as adaptation models for evaluation after the training is completed. A different training paradigm from regular Co-Teaching is adopted, which utilizes a different mechanism to filter and acquire low-loss samples. It enables the network to counter noisy correspondence from different perspectives and improves model robustness. Moreover, this approach prevents assimilation between the two models in Co-Teaching, avoiding negative impacts on prediction effectiveness.

\begin{table*}[ht]
  \centering
  \resizebox{\textwidth}{!}{%
    \begin{tabular}{c|c|ccccccc|ccccccc}
    \hline
    \multirow{3}{*}{Noise} & \multirow{3}{*}{Methods} & \multicolumn{7}{c|}{Flick30K} & \multicolumn{7}{c}{MSCOCO} \\ \cline{3-16} 
    & & \multicolumn{3}{c|}{Image-Text} & \multicolumn{3}{c|}{Text-Image} & \multirow{2}{*}{Rsum} & \multicolumn{3}{c|}{Image-Text} & \multicolumn{3}{c|}{Text-Image} & \multirow{2}{*}{Rsum} \\ \cline{3-8} \cline{10-15}
    & & R@1 & R@5 & \multicolumn{1}{c|}{R@10} & R@1 & R@5 & \multicolumn{1}{c|}{R@10} & & R@1 & R@5 & \multicolumn{1}{c|}{R@10} & R@1 & R@5 & \multicolumn{1}{c|}{R@10} & \\ \hline
    \multirow{11}{*}{20\%} & SCAN & 58.5 & 81.0 & \multicolumn{1}{c|}{90.8} & 35.5 & 65.0 & \multicolumn{1}{c|}{75.2} & 406.0 & 62.2 & 90.0 & \multicolumn{1}{c|}{96.1} & 46.2 & 80.8 & \multicolumn{1}{c|}{89.2} & 464.5 \\
    & IMRAM & 22.7 & 54.0 & \multicolumn{1}{c|}{67.8} & 16.6 & 41.8 & \multicolumn{1}{c|}{54.1} & 257.0 & 69.9 & 93.6 & \multicolumn{1}{c|}{97.4} & 55.9 & 84.4 & \multicolumn{1}{c|}{89.6} & 490.8 \\
    & SAF & 62.8 & 88.7 & \multicolumn{1}{c|}{93.9} & 49.7 & 73.6 & \multicolumn{1}{c|}{78.0} & 446.7 & 71.5 & 94.0 & \multicolumn{1}{c|}{97.5} & 57.8 & 86.4 & \multicolumn{1}{c|}{91.9} & 499.1 \\
    & SGR & 55.9 & 81.5 & \multicolumn{1}{c|}{88.9} & 40.2 & 66.8 & \multicolumn{1}{c|}{75.3} & 408.6 & 25.7 & 58.8 & \multicolumn{1}{c|}{75.1} & 23.5 & 58.9 & \multicolumn{1}{c|}{75.1} & 317.1 \\
    & NCR* & 73.5 & 93.2 & \multicolumn{1}{c|}{96.6} & 56.9 & 82.4 & \multicolumn{1}{c|}{88.5} & 491.1 & 76.6 & 95.6 & \multicolumn{1}{c|}{98.2} & 60.8 & 88.8 & \multicolumn{1}{c|}{95.0} & 515.0 \\
    & MSCN* & 77.4 & \underline{94.9} & \multicolumn{1}{c|}{\underline{97.6}} & 59.6 & 83.2 & \multicolumn{1}{c|}{89.2} & 501.9 & 78.1 & \underline{97.2} & \multicolumn{1}{c|}{\underline{98.8}} & 64.3 & 90.4 & \multicolumn{1}{c|}{95.8} & 524.6 \\
    & BiCro* & 78.1 & 94.4 & \multicolumn{1}{c|}{97.5} & \underline{60.4} & \textbf{84.4} & \multicolumn{1}{c|}{\underline{89.9}} & \underline{504.7} & 78.8 & 96.1 & \multicolumn{1}{c|}{98.6} & 63.7 & 90.3 & \multicolumn{1}{c|}{95.7} & 523.2 \\
    & ESC* & \underline{79.0} & 94.8 & \multicolumn{1}{c|}{97.5} & 59.1 & 83.8 & \multicolumn{1}{c|}{89.1} & 503.3 & \underline{79.2} & 97.0 & \multicolumn{1}{c|}{\textbf{99.1}} & \underline{64.8} & \underline{90.7} & \multicolumn{1}{c|}{\underline{96.0}} & \underline{526.8} \\
    & \textbf{TSVC(Ours)} & \textbf{79.6} & \textbf{95.7} & \multicolumn{1}{c|}{\textbf{98.6}} & \textbf{60.9} & \underline{84.1} & \multicolumn{1}{c|}{\textbf{90.9}} & \textbf{509.8} & \textbf{79.9} & \textbf{97.5} & \multicolumn{1}{c|}{98.6} & \textbf{65.3} & \textbf{91.8} & \multicolumn{1}{c|}{\textbf{97.3}} & \textbf{530.4} \\ \hline
    \multirow{11}{*}{40\%} & SCAN & 26.0 & 57.4 & \multicolumn{1}{c|}{71.8} & 17.8 & 40.5 & \multicolumn{1}{c|}{51.4} & 264.9 & 42.9 & 74.6 & \multicolumn{1}{c|}{85.1} & 24.2 & 52.6 & \multicolumn{1}{c|}{63.8} & 343.2 \\
    & IMRAM & 5.3 & 25.4 & \multicolumn{1}{c|}{37.6} & 5.0 & 13.5 & \multicolumn{1}{c|}{19.6} & 106.4 & 51.8 & 82.4 & \multicolumn{1}{c|}{90.9} & 38.4 & 70.3 & \multicolumn{1}{c|}{78.9} & 412.7 \\
    & SAF & 7.4 & 19.6 & \multicolumn{1}{c|}{26.7} & 4.4 & 12.2 & \multicolumn{1}{c|}{17.0} & 87.3 & 13.5 & 43.8 & \multicolumn{1}{c|}{48.2} & 16.0 & 39.0 & \multicolumn{1}{c|}{50.8} & 211.3 \\
    & SGR & 4.1 & 16.6 & \multicolumn{1}{c|}{24.1} & 4.1 & 13.2 & \multicolumn{1}{c|}{19.7} & 81.8 & 1.3 & 3.7 & \multicolumn{1}{c|}{6.3} & 0.5 & 2.5 & \multicolumn{1}{c|}{4.1} & 18.4 \\
    & NCR* & 75.3 & 92.1 & \multicolumn{1}{c|}{95.2} & 56.2 & 80.6 & \multicolumn{1}{c|}{\underline{87.4}} & 486.8 & 76.5 & 95.0 & \multicolumn{1}{c|}{98.2} & 60.7 & 88.5 & \multicolumn{1}{c|}{95.0} & 513.9 \\
    & MSCN* & 74.4 & \underline{94.4} & \multicolumn{1}{c|}{\underline{96.9}} & \underline{57.2} & \underline{81.7} & \multicolumn{1}{c|}{\textbf{87.6}} & \underline{492.2} & 74.8 & 94.9 & \multicolumn{1}{c|}{98.0} & 60.3 & 88.5 & \multicolumn{1}{c|}{94.4} & 510.9 \\
    & BiCro* & 74.6 & 92.7 & \multicolumn{1}{c|}{96.2} & 55.5 & 81.1 & \multicolumn{1}{c|}{\underline{87.4}} & 487.5 & 77.0 & 95.9 & \multicolumn{1}{c|}{98.3} & 61.8 & 89.2 & \multicolumn{1}{c|}{94.9} & 517.1 \\
    & ESC* & \underline{76.1} & 93.1 & \multicolumn{1}{c|}{96.4} & 56.0 & 80.8 & \multicolumn{1}{c|}{87.2} & 489.6 & \underline{78.6} & \underline{96.6} & \multicolumn{1}{c|}{\textbf{99.0}} & \underline{63.2} & \underline{90.6} & \multicolumn{1}{c|}{\underline{95.9}} & \underline{523.9} \\
    & \textbf{TSVC(Ours)} & \textbf{77.7} & \textbf{95.3} & \multicolumn{1}{c|}{\textbf{98.3}} & \textbf{58.8} & \textbf{83.1} & \multicolumn{1}{c|}{87.1} & \textbf{500.3} & \textbf{79.0} & \textbf{97.3} & \multicolumn{1}{c|}{\underline{98.6}} & \textbf{65.5} & \textbf{91.3} & \multicolumn{1}{c|}{\textbf{96.2}} & \textbf{527.9} \\ \hline
    \multirow{11}{*}{60\%} & SCAN & 13.6 & 36.5 & \multicolumn{1}{c|}{50.3} & 4.8 & 13.6 & \multicolumn{1}{c|}{19.8} & 138.6 & 29.9 & 60.9 & \multicolumn{1}{c|}{74.8} & 0.9 & 2.4 & \multicolumn{1}{c|}{4.1} & 173.0 \\
    & IMRAM & 1.5 & 8.9 & \multicolumn{1}{c|}{17.4} & 1.9 & 5.0 & \multicolumn{1}{c|}{7.8} & 42.5 & 18.2 & 51.6 & \multicolumn{1}{c|}{68.0} & 17.9 & 43.6 & \multicolumn{1}{c|}{54.6} & 253.9 \\
    & SAF & 0.1 & 1.5 & \multicolumn{1}{c|}{2.8} & 0.4 & 1.2 & \multicolumn{1}{c|}{2.3} & 8.3 & 0.1 & 0.5 & \multicolumn{1}{c|}{0.7} & 0.8 & 3.5 & \multicolumn{1}{c|}{6.3} & 11.9 \\
    & SGR & 1.5 & 6.6 & \multicolumn{1}{c|}{9.6} & 0.3 & 2.3 & \multicolumn{1}{c|}{4.2} & 24.5 & 0.1 & 0.6 & \multicolumn{1}{c|}{1.0} & 0.1 & 0.5 & \multicolumn{1}{c|}{1.1} & 3.4 \\
    & NCR* & 13.9 & 37.7 & \multicolumn{1}{c|}{50.5} & 11.0 & 30.1 & \multicolumn{1}{c|}{41.4} & 184.6 & 0.1 & 0.3 & \multicolumn{1}{c|}{0.4} & 0.1 & 0.5 & \multicolumn{1}{c|}{1.0} & 2.4 \\
    & MSCN* & 70.4 & \underline{91.0} & \multicolumn{1}{c|}{\underline{94.9}} & \underline{53.4} & 77.8 & \multicolumn{1}{c|}{84.1} & 471.6 & 74.4 & \underline{95.1} & \multicolumn{1}{c|}{97.9} & 59.2 & 87.1 & \multicolumn{1}{c|}{92.8} & 506.5 \\
    & BiCro* & 67.6 & 90.8 & \multicolumn{1}{c|}{94.4} & 51.2 & 77.6 & \multicolumn{1}{c|}{84.7} & 466.3 & 73.9 & 94.4 & \multicolumn{1}{c|}{97.8} & 58.3 & 87.2 & \multicolumn{1}{c|}{\underline{93.9}} & 505.5 \\
    & ESC* & \underline{72.6} & 90.9 & \multicolumn{1}{c|}{94.6} & 53.0 & \textbf{78.6} & \multicolumn{1}{c|}{\underline{85.3}} & \underline{475.0} & \underline{77.2} & \underline{95.1} & \multicolumn{1}{c|}{\underline{98.1}} & \underline{61.1} & \underline{88.6} & \multicolumn{1}{c|}{\textbf{94.9}} & \underline{515.0} \\
    & \textbf{TSVC(Ours) }& \textbf{73.2} & \textbf{92.0} & \multicolumn{1}{c|}{\textbf{95.1}} & \textbf{54.8} & \underline{78.5} & \multicolumn{1}{c|}{\textbf{86.2}} & \textbf{479.8} & \textbf{77.4} & \textbf{96.8} & \multicolumn{1}{c|}{\textbf{99.5}} & \textbf{61.7} & \textbf{89.0} & \multicolumn{1}{c|}{\textbf{94.9}} & \textbf{519.3} \\ \hline
    \end{tabular}
  }
  \caption{Image-Text Retrieval on Flickr30K and MS-COCO 1K datasets under different noise ratios. * indicates the noise robust method. The best and sub-optimal indicators are represented in \textbf{bold} and \underline{underline} respectively.}
  \label{table1}
\end{table*}

\begin{figure*}[htbp]
    \centering    

    \begin{minipage}{.22\textwidth}  
        \includegraphics[width=\linewidth]{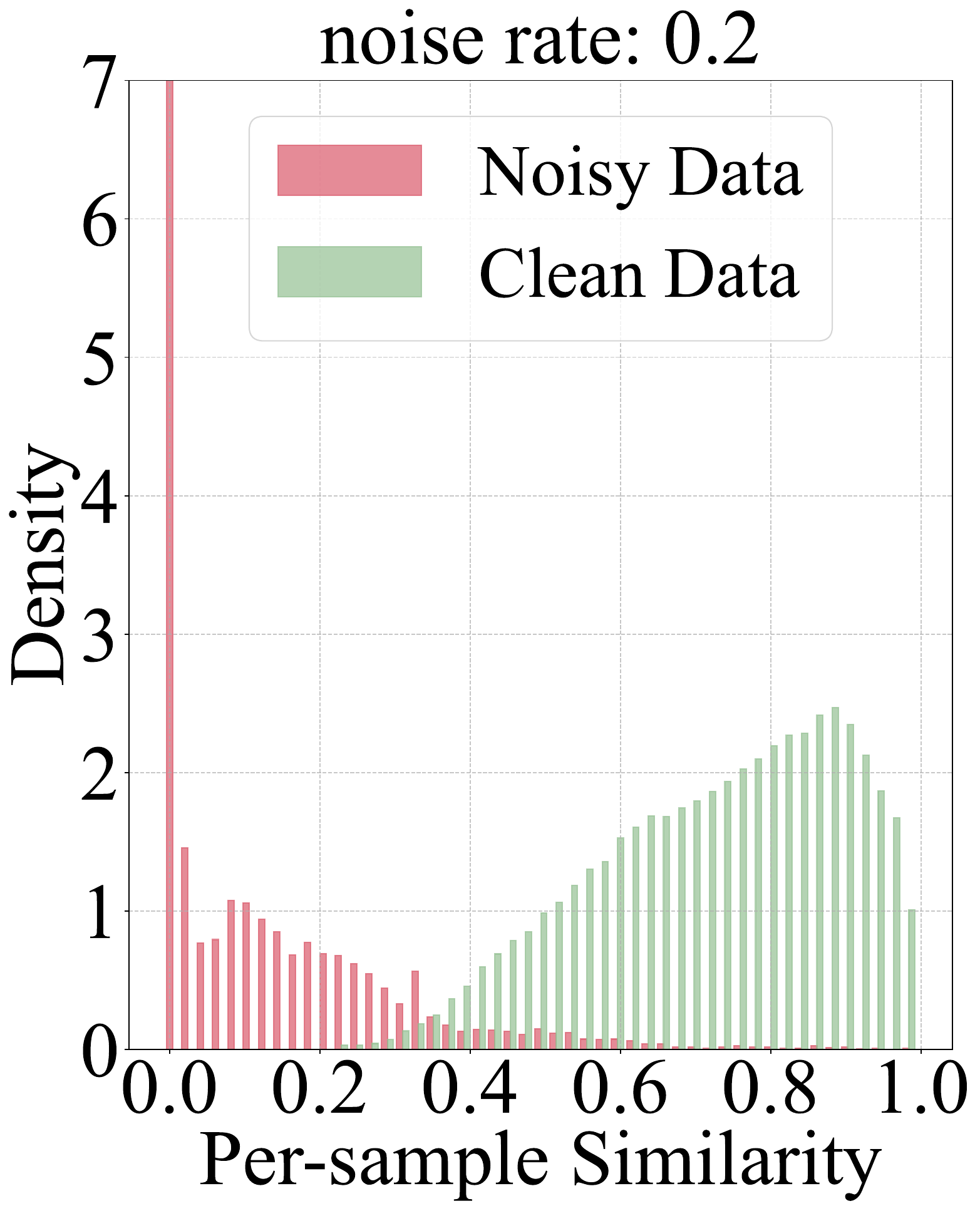}
        \centering
        \subcaption{\hspace{5pt}W/o DASM}
        \label{fig:sub1}  
    \end{minipage}
    \hfill
    \begin{minipage}{.22\textwidth}  
        \includegraphics[width=\linewidth]{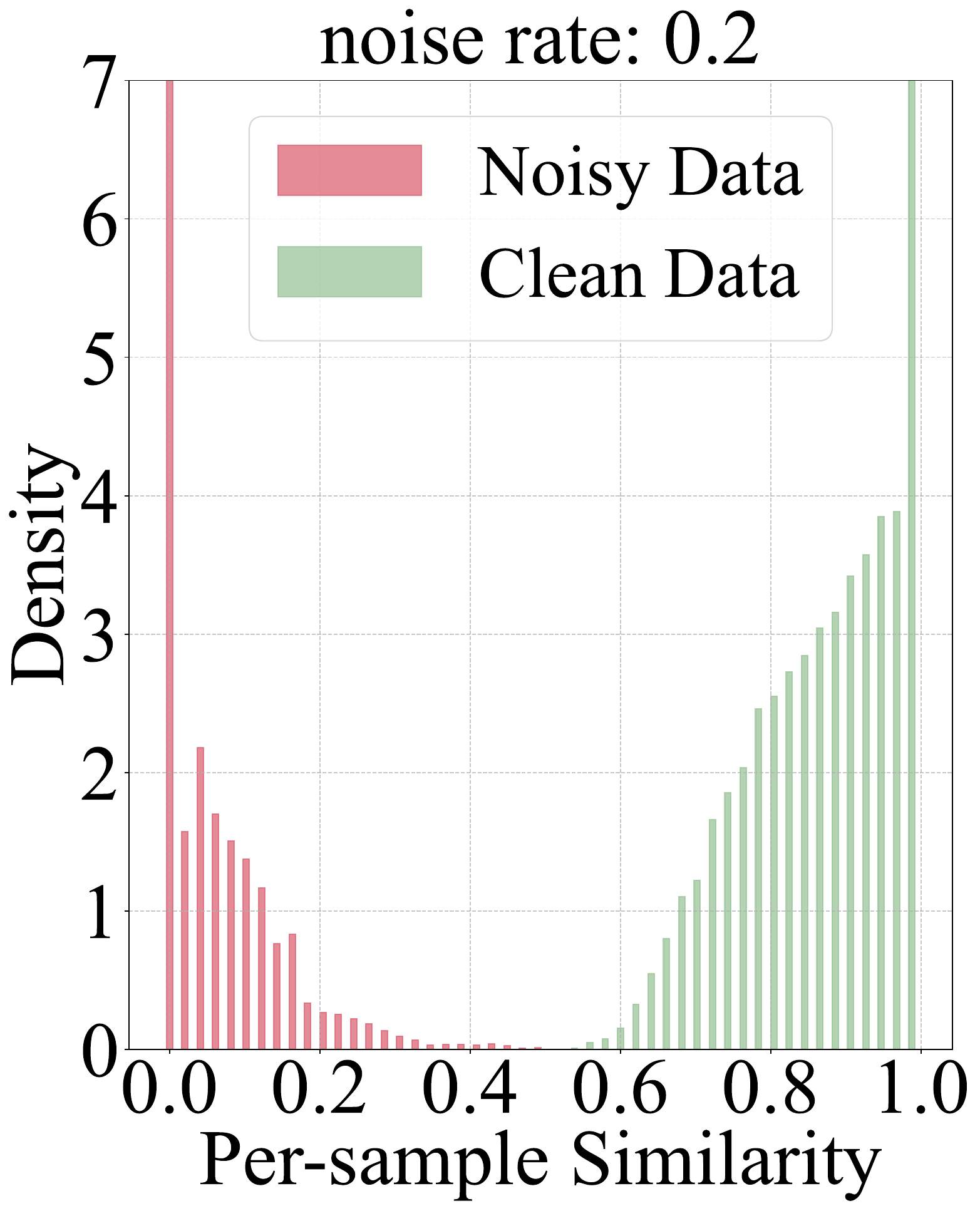}
        \centering
        \subcaption{\hspace{5pt}W/ DASM}
        \label{fig:sub2}  
    \end{minipage}
    \hfill
    \begin{minipage}{.22\textwidth}  
        \includegraphics[width=\linewidth]{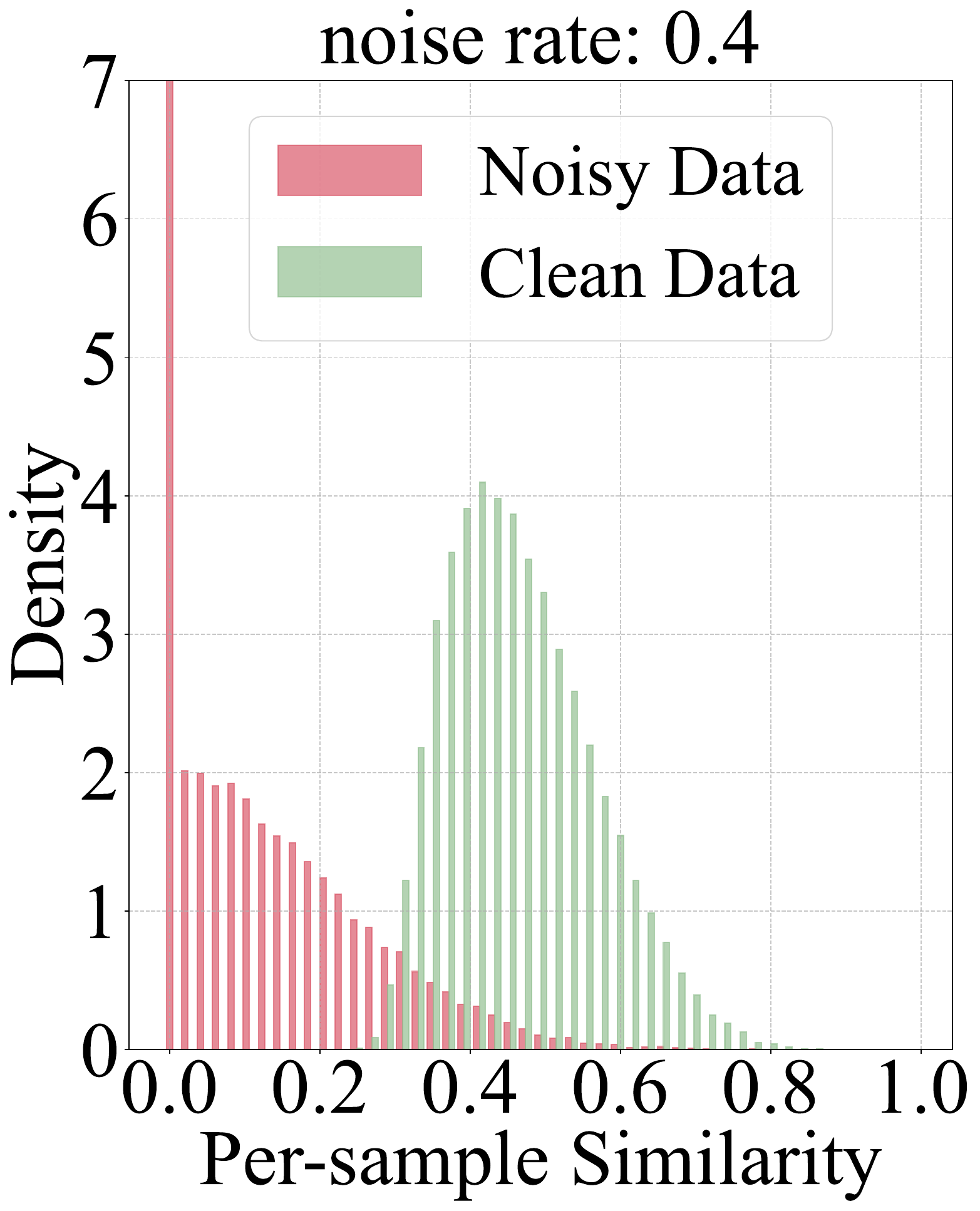}
        \centering
        \subcaption{\hspace{5pt}W/o DASM}
        \label{fig:sub3}  
    \end{minipage}
    \hfill
    \begin{minipage}{.22\textwidth}  
        \includegraphics[width=\linewidth]{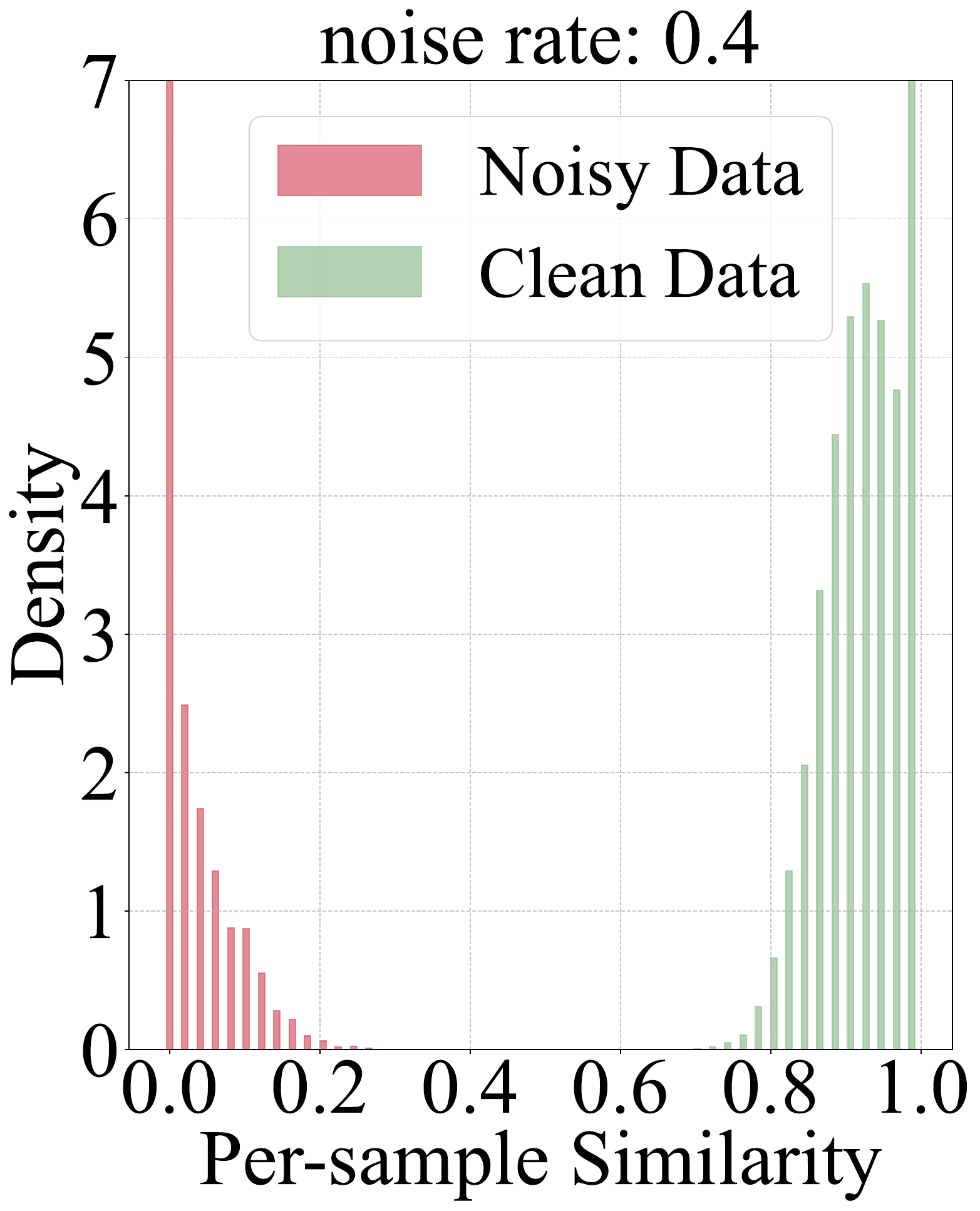}
        \centering
        \subcaption{\hspace{5pt}W/ DASM}
        \label{fig:sub4}  
    \end{minipage}

    \caption{Distribution change on Flickr30K with noise ratios of 20\% and 40\%.} 
    \label{fig:dasm}
    \label{topk}  
\end{figure*}

\begin{figure}[t]
    \centering    

    \begin{minipage}{.22\textwidth}  
        \includegraphics[width=\linewidth]{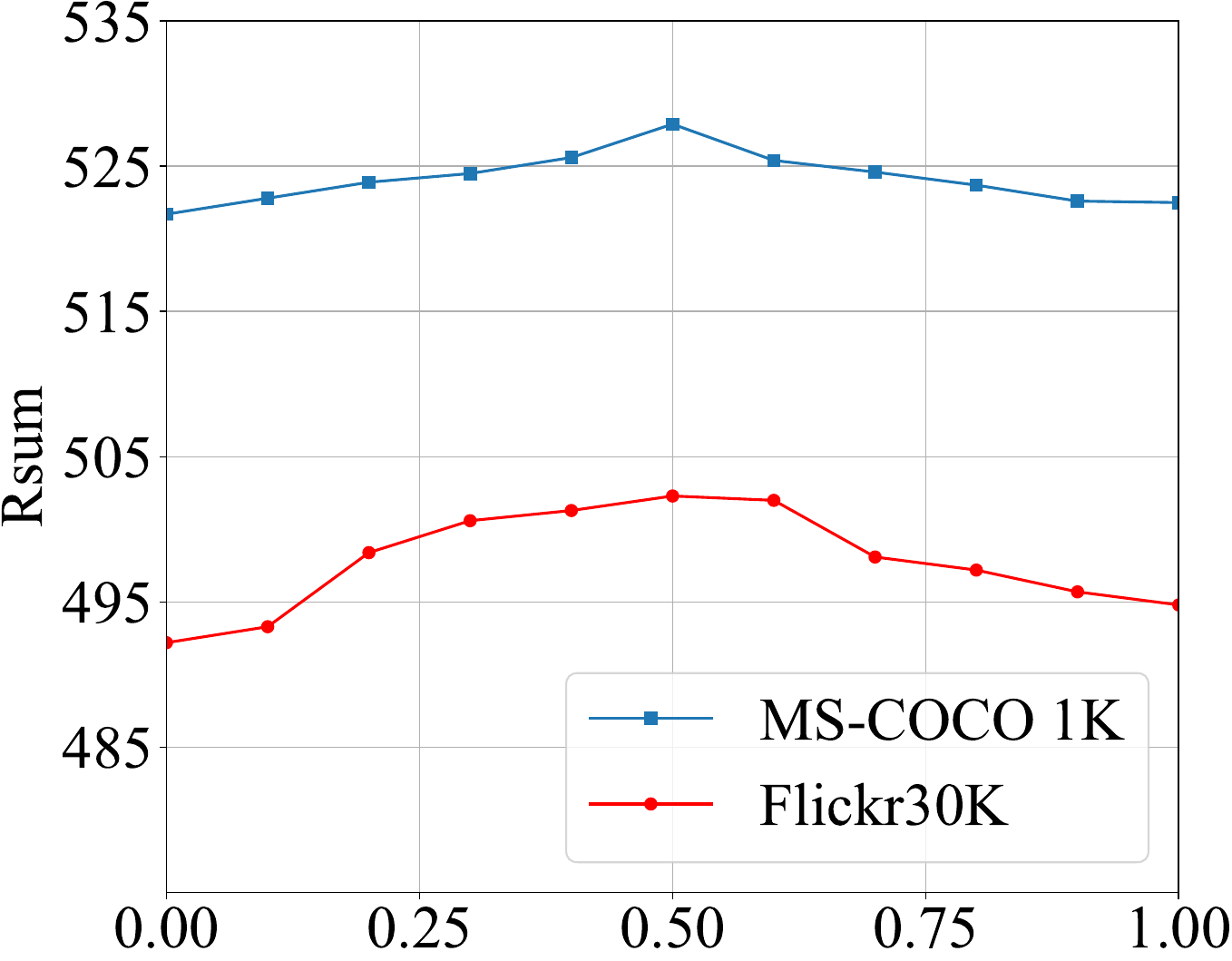}
        \centering
        \subcaption{\hspace{5pt}$\delta$}
        \label{fig:mismatch_threshold}  
    \end{minipage}
    \hfill
    \begin{minipage}{.22\textwidth}  
        \includegraphics[width=\linewidth]{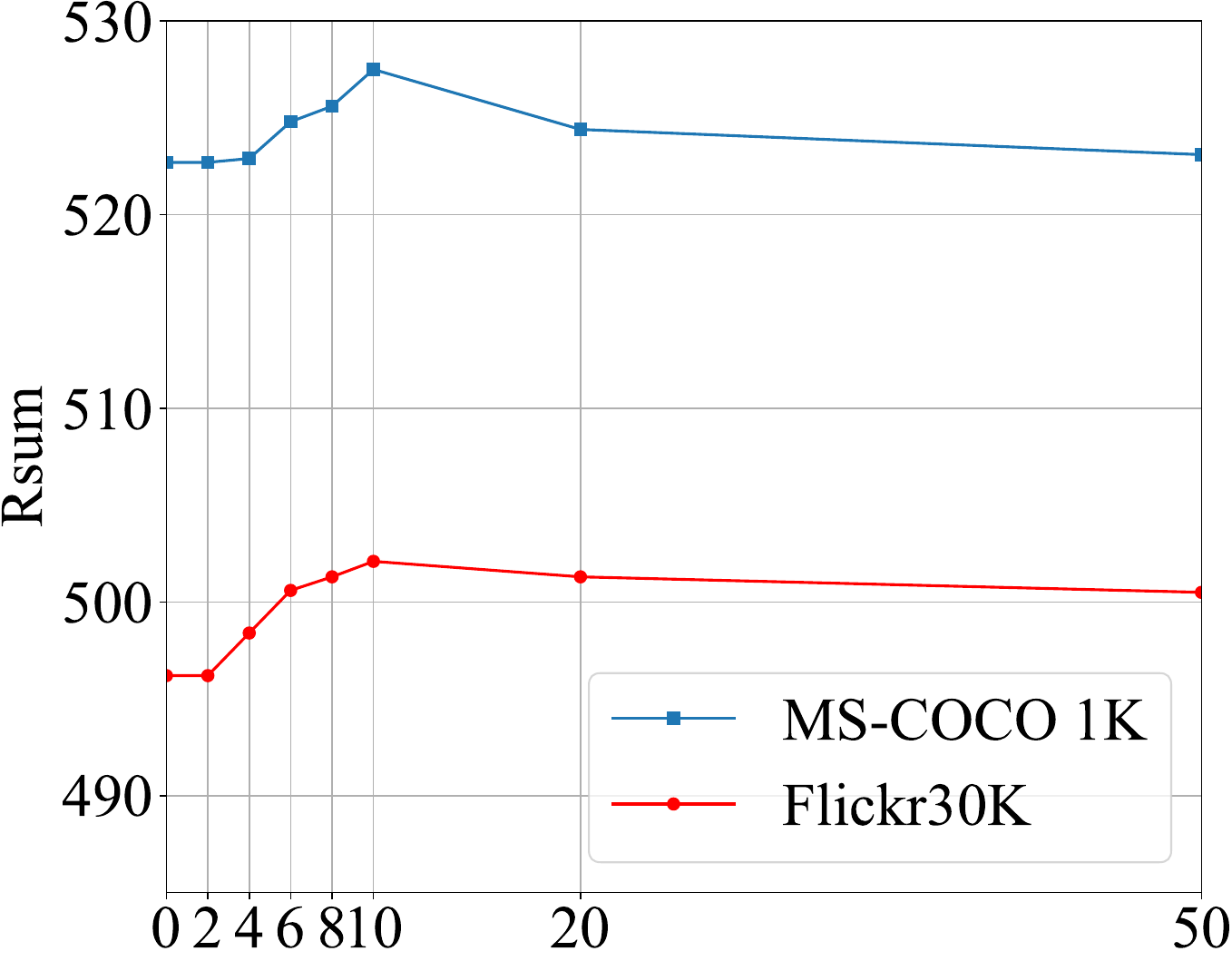}  
        \centering
        \subcaption{\hspace{5pt}$m$}
        \label{fig:hyperparameter_m}  
    \end{minipage}

     \caption{Analysis of hyper-parameters $\delta$ and $m$ on Flickr30K with 40\% noise. \textbf{Left:} $\delta$ is the filtering threshold. \textbf{Right:} $m$ is the parameter of the soft-margin formula. } 
     \label{fig:hyperparameters_analysis}
\end{figure}

\subsection{Distribution-Adaptive Soft Margin Loss}


We propose DASM loss, which adjusts the soft margin based on the similarity between sample pairs and their deviation from the center of the clean distribution. Specifically, for an image-text pair $(I', T')$ in a mini-batch, we compute the distance from it to the center of the clean sample loss distribution $d=|L-L_{clean}|$, and then calculate the loss $\mathcal{L}_{DASM}$:
\begin{equation}
    \begin{aligned}
        \mathcal{L}_{DASM}(I_i,T_i)=&\left[ \tilde{a}_i-S\left( I_i,T_i \right) +S\left( I_i,\hat{T}_n \right) \right] _+ \\
      &  +\left[ \tilde{a}_i-S\left( I_i,T_i \right)  
        +S\left( 
	\hat{I}_{n},T_{i} 
\right) \right] _+.
    \end{aligned}
\end{equation}
\begin{equation}
    \tilde{a}_i=(2+tanh(-d))\frac{m^{y^*_i} - 1}{m-1}\alpha,
\end{equation}

\begin{table}[t]
  \centering

  \setlength{\tabcolsep}{1mm} 
  \resizebox{\columnwidth}{!}{%
  \begin{tabular}{c|ccccccc}
    \hline
    \multirow{3}{*}{Methods} & \multicolumn{7}{c}{CC152K}                                                                                \\ \cline{2-8} 
                             & \multicolumn{3}{c|}{Image-Text}         & \multicolumn{3}{c|}{Text-Image}         & \multirow{2}{*}{Rsum} \\ \cline{2-7}
                             & R@1  & R@5  & \multicolumn{1}{c|}{R@10} & R@1  & R@5  & \multicolumn{1}{c|}{R@10} &                       \\ \hline
    SCAN                     & 30.5 & 55.3 & \multicolumn{1}{c|}{65.3} & 26.9 & 53.0 & \multicolumn{1}{c|}{64.7} & 295.7                 \\
    IMRAM                    & 33.1 & 57.6 & \multicolumn{1}{c|}{68.1} & 29.0 & 56.8 & \multicolumn{1}{c|}{67.4} & 312.0                 \\
    SAF                      & 31.7 & 59.3 & \multicolumn{1}{c|}{68.2} & 31.9 & 59.0 & \multicolumn{1}{c|}{67.9} & 318.0                 \\
    SGR                      & 11.3 & 29.7 & \multicolumn{1}{c|}{39.6} & 13.1 & 30.1 & \multicolumn{1}{c|}{41.6} & 165.4                 \\
    NCR*                     & 39.5 & 64.5 & \multicolumn{1}{c|}{73.5} & 40.3 & 64.6 & \multicolumn{1}{c|}{73.2} & 355.6                 \\
    MSCN*                    & 40.1 & 65.7 & \multicolumn{1}{c|}{\underline{76.6}} & 40.6 & 67.4 & \multicolumn{1}{c|}{76.3} & 366.7                 \\
    BiCro*                   & 40.8 & 67.2 & \multicolumn{1}{c|}{76.1} & 42.1 & 67.6 & \multicolumn{1}{c|}{\underline{76.4}} & 370.2                 \\
    ESC*                     & \underline{42.8} & \underline{67.3} & \multicolumn{1}{c|}{\textbf{76.9}} & \underline{44.8} & \textbf{68.2} & \multicolumn{1}{c|}{75.9} & \underline{375.9} \\
    \textbf{TSVC(Ours)}                      & \textbf{44.7} & \textbf{69.4} & \multicolumn{1}{c|}{76.5} & \textbf{45.1} & \underline{67.8} & \multicolumn{1}{c|}{\textbf{76.6}} & \textbf{380.1} \\ \hline
  \end{tabular}%
  }
\caption{Image-Text Retrieval on CC152K. * indicates the noise robust method. The best and sub-optimal indicators are represented in \textbf{bold} and \underline{underline} respectively.}
\label{table2}
\end{table}

where $\hat{T}_n=argmax_{T_j\neq T_i} S(I_i,T_j)$ and $\hat{I}_n =argmax_{I_j\neq I_i} S(I_j,T_i)$ represent the hard negative text and image that are most similar to the aligned image-text pair $(I_i,T_i)$ within a batch. As iterations progress, DASM gradually redirects samples that significantly deviate from the clean distribution center, strengthening the boundary between clean and noisy distributions. By slowing down the boundary growth deliberately, it avoids overly penalizing misclassification similarity scores, thus maintaining model balance and flexibility.

\section{Experiments }
\subsection{Datasets}
We utilize the following three widely used multi-modal datasets to evaluate our method: 

\noindent
\textbf{Flickr30K} This dataset contains 31,000 images collected from Flickr, with each image paired with five textual descriptions providing detailed annotations of the content. It covers a variety of scenes, objects, and activities, making it a common benchmark for cross-modal retrieval tasks. In our experiments, we use 29,000 images for training, 1,000 for validation, and 1,000 for testing to evaluate the model's performance.

\noindent
\textbf{MSCOCO} This dataset consists of 123,287 images, each annotated with five textual captions describing diverse visual content, including daily scenes and objects. It serves as a critical benchmark for multimodal tasks. For our experiments, 113,287 images are used for training, 5,000 for validation, and 5,000 for testing, ensuring a comprehensive performance evaluation.

\noindent
\textbf{Conceptual Captions.} A large scale dataset mainly collected from the Internet, which comprises 3.3 million pictures, each accompanied by a corresponding caption. Around 3\%-20\% of the image-text pairs in the dataset exhibit inconsistencies~\cite{sharma2018conceptual}. Similar to previous studies~\cite{huang2021learning}, we use the CC152K subset from Conceptual Captions in our experiments, which contains 150k images for training, and 1000 images each for validation and testing.

\subsection{Evaluation Metrics}
We employ Recall@K (R@K) to measure the retrieval performance, which quantifies the accuracy of identifying relevant items within the top K results. Additionally, to provide a more comprehensive analysis, we report R@1, R@5, and R@10 for both image-to-text and text-to-image retrieval tasks, and summarize these metrics as Rsum for a comprehensive performance evaluation.

\subsection{Main Results}
In this section, we validate performance of the TSVC network on the aforementioned three datasets. Specifically, we select 4 methods without noise learning, i.e., SCAN~\cite{lee2018stacked}, IMRAM~\cite{chen2020imram}, SAF and SGR~\cite{diao2021similarity}, to illustrate the impact brought by noise. We select another 4 methods with noise learning, i.e., NCR~\cite{huang2021learning}, MSCN~\cite{han2023noisy}, BiCro~\cite{yang2023bicro}, and ESC~\cite{yang2024robust}), to exhibit superior performance of TSVC. 

\noindent\textbf{Experiments on Simulated Noise.} To evaluate performance under noise, we introduce simulated noise by randomly shuffling image-caption pairs in Flickr30K and MS-COCO, with noise ratios of 20\%, 40\%, and 60\%. Results in Table~\ref{table1} show that our proposed TSVC method consistently outperforms state-of-the-art methods. Compared to the best-performing baseline \textbf{ESC}~\cite{yang2024robust}, TSVC achieves improvements of 6.5, 10.7, and 4.8 on Flickr30K, and 3.6, 4.0, and 4.3 on MSCOCO-1k under the respective noise ratios.

\noindent\textbf{Experiments on Real-world Noise.} We evaluate our method on the real-world noisy scenes in the CC152K dataset. It includes mismatched pairs with unknown parts and serves as a challenging benchmark dataset. Detailed experimental results are listed in Table~\ref{table2}. From the table, we can observe that even under the condition of real noise, TSVC still demonstrates competitive performance. Specifically, TSVC outperforms the state-of-the-art baseline method ESC by 4.2 for overall retrieval performance. It indicates the effectiveness of TSVC in improving retrieval accuracy.

\begin{table}[t]
  \centering
  \setlength{\tabcolsep}{1.1mm} 
  \renewcommand{\arraystretch}{1.2}
  \resizebox{0.45\textwidth}{!}{ %
    \begin{tabular}{c|c|c|ccc|ccc}
      \hline
      \multicolumn{3}{c|}{\multirow{2}{*}{Configuration}} & \multicolumn{3}{c|}{Image-Text}               & \multicolumn{3}{c}{Text-Image}                \\ 
      \cline{4-9} 
      \multicolumn{3}{c|}{}                               & R@1           & R@5           & R@10          & R@1           & R@5           & R@10          \\ \hline
      \multicolumn{3}{c|}{\textbf{TSVC}}                           & \textbf{77.7} & \textbf{95.3} & \textbf{98.3} & \textbf{58.8} & \textbf{83.1} & \textbf{87.1} \\
      \multicolumn{3}{c|}{w/o SIVC}                 & 74.8          & 93.6          & 96.7          & 55.4          & 80.2          & 84.3          \\
      \multicolumn{3}{c|}{w/o DASM}                       & 75.6          & 94.1          & 97.0          & 57.2          & 81.5          & 85.8          \\
      \multicolumn{3}{c|}{w/o Tri-Learning}               & 75.0          & 94.3          & 97.7          & 58.0          & 81.1          & 85.7          \\ \hline
    \end{tabular}
  }
    \caption{Ablation studies on Flickr30K with 40\% noise ratio.}
    \label{ablation}
\end{table}
 \subsection{Ablation Study}
We conduct an ablation study on three datasets to investigate the impact of individual components on performance. Due to space limit, we only present results on the Flickr30K dataset with 40\% noise impact. The results are listed in Table~\ref{ablation}. Similar trends are observed across the other datasets. We remove three key components of \textbf{TSVC}, i.e., SIVC, DASM, and Tri-learning, to observe their impacts.
 
To verify the effectiveness of SIVC, we employ hard labels for comparison, where clean samples are labeled as 1 and noisy samples as 0. To test the effectiveness of the distance metric DASM, we replace it with a regular soft-margin triplet loss function. To verify the effectiveness of the Tri-learning scheme, we compare it with the Co-teaching approach.
 
 Results are presented in Table~\ref{ablation}. From the table, we can make three observations. (1) Employing SIVC for soft label estimation outperforms relying solely on hard labels. It suggests that soft labels provide better measurement of matching degree between data pairs and help the models learn useful information. (2) The use of distance metric loss function is significantly superior to using ordinary triplet loss. Meanwhile, we visualize the sample similarity distribution before and after using DASM. As shown in Fig.~\ref{fig:dasm}, the improvement in data partitioning is significant. (3) The Tri-learning architecture demonstrates superior performance compared to the Co-training paradigm. This is because Tri-learning enables each model within the architecture to fulfill its designated role and collaborate with others. Thus, it effectively mitigates model homogenization and enhances network performance. (4) The complete TSVC model achieves the best overall performance, demonstrating the importance of all three components in terms of noise correspondence.
 \begin{figure}[t]
    \centering    %
    \begin{minipage}{\linewidth}  
        \includegraphics[width=\linewidth]{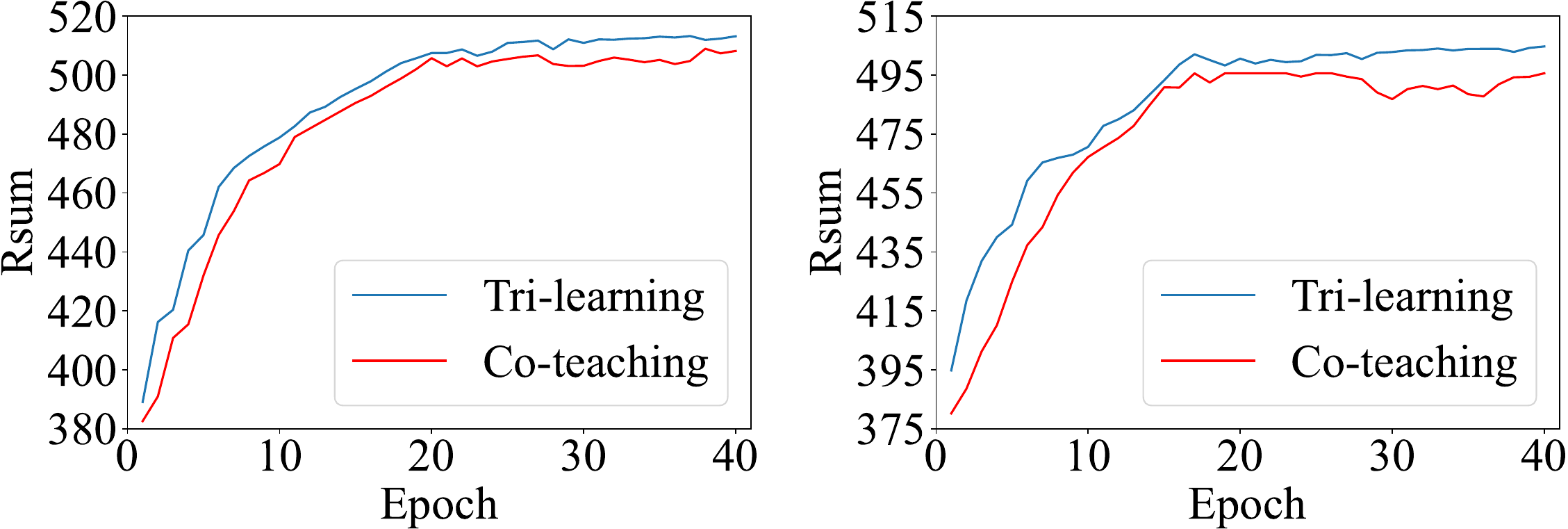}
        \centering
        \subcaption{noise ratio: 0 (Left) and 0.2 (Right).}
        \label{fig:tri_02}  
    \end{minipage}
    \begin{minipage}{\linewidth}  
        \includegraphics[width=\linewidth]{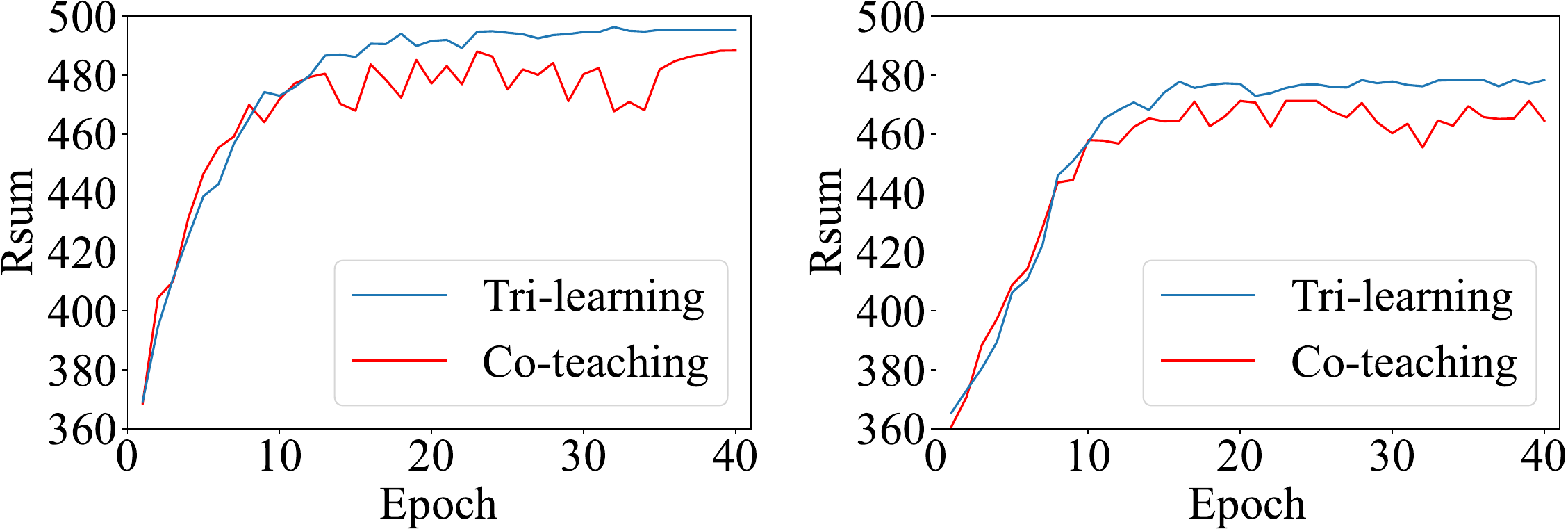}
        \centering
        \subcaption{noise ratio: 0.4 (Left) and 0.6 (Right).}
        \label{fig:tri_46}  
    \end{minipage}

    \caption{Results on Flickr30K dataset.} 
    \label{fig:learning_curve}
\end{figure}

\subsection{Analysis of the Tri-learning Method}
Figure~\ref{fig:learning_curve} illustrates the test Rsum values during the training process under different noise types. We can observe that when the noise ratio is 0.2, the learning curves of both Co-teaching and Tri-learning are relatively stable. 
Nevertheless, as the noise ratio increases, the learning curve of Co-teaching exhibits greater fluctuations. Because of homogenization of the two models, the final predictive performance of Co-teaching is limited. 
In contrast, the learning curve of Tri-learning tends to fit the data more easily, exhibiting greater stability. This indicates that Tri-learning is more effective in handling higher levels of noise, achieving more robust performance in noisy environment. 

Notably, as shown in Fig.~\ref{fig:learning_curve}\,(b), under high noise ratios, Tri-learning initially performs worse than Co-teaching during training, because it may misclassify noisy samples as clean samples in the Re-Division stage. Nevertheless, as it is fully trained, its performance demonstrates greater stability than that of Co-teaching.

\subsection{Hyperparameters Analysis}
The hyperparameters $\delta$ and $m$ represent the threshold used for dividing noisy samples and the parameter controlling the soft margin in DASM, respectively. These parameters play a important role in balancing noise filtering and optimizing margin flexibility. We conduct experiments on Flickr30K and MSCOCO datasets with a noise ratio of 40\%.  As depicted in Fig.~\ref{fig:hyperparameters_analysis}, the value of Rsum increases continuously with an increase in $\delta$, reaching its peak at $\delta=0.5$ (between 0.5 to 0.6 for MSCOCO), and then gradually decreases. A larger mismatch threshold $\delta$ might classify weakly labeled samples as mislabeled ones, thus impairing generalization ability. The optimal value for $m$ is 10, which significantly influences the performance of the model by affecting the size of the soft-margin. Therefore, it should be considered comprehensively during the model optimization process to ensure the best performance.

\begin{figure}[t]
    \centering
    \includegraphics[width=8.4cm]{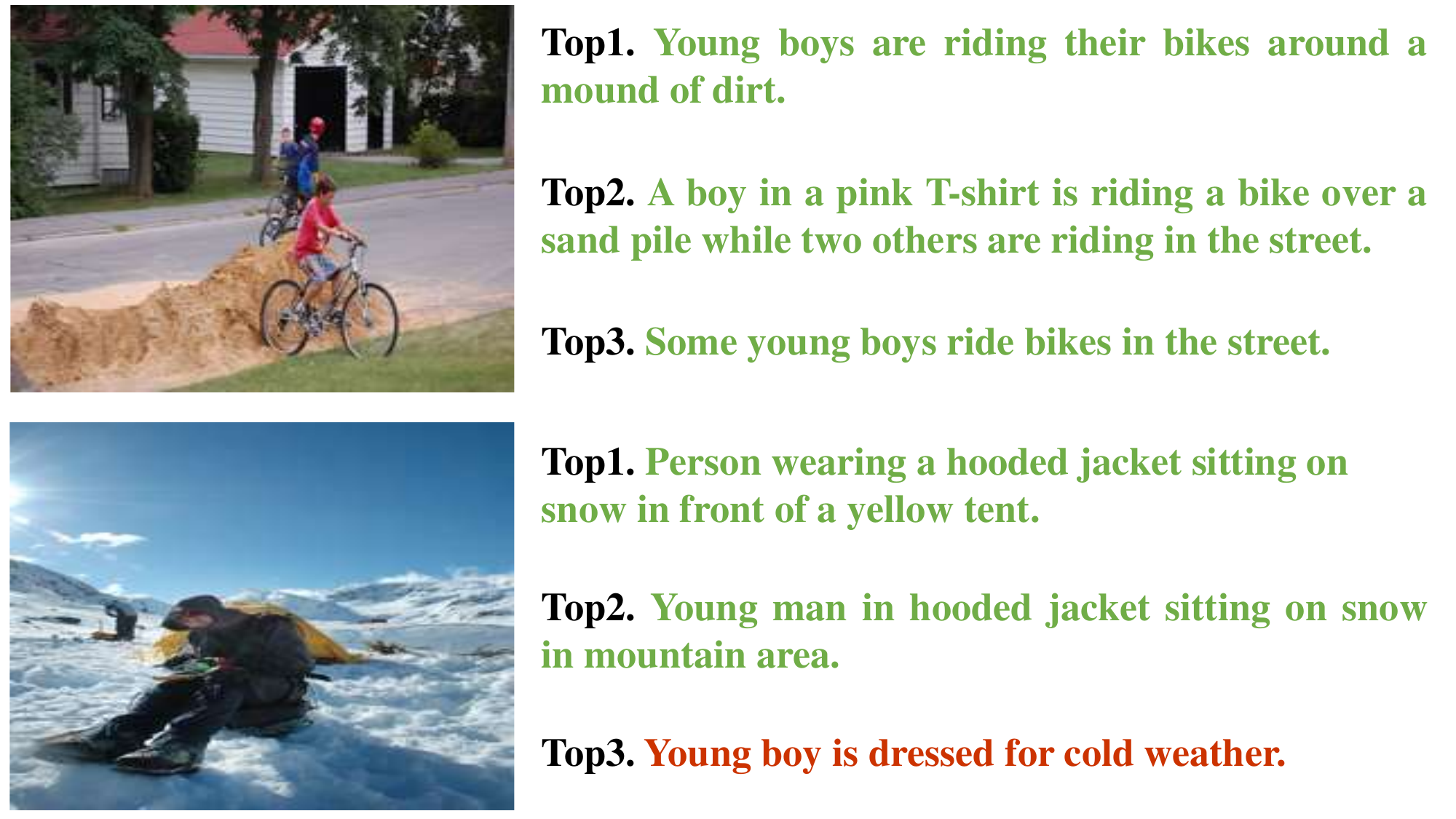}
    \caption{Image-To-Text matching results on Flickr30K.}
    \label{fig:case_1}
\end{figure}

\begin{figure}[t]
    \centering
    \includegraphics[width=8.4cm]{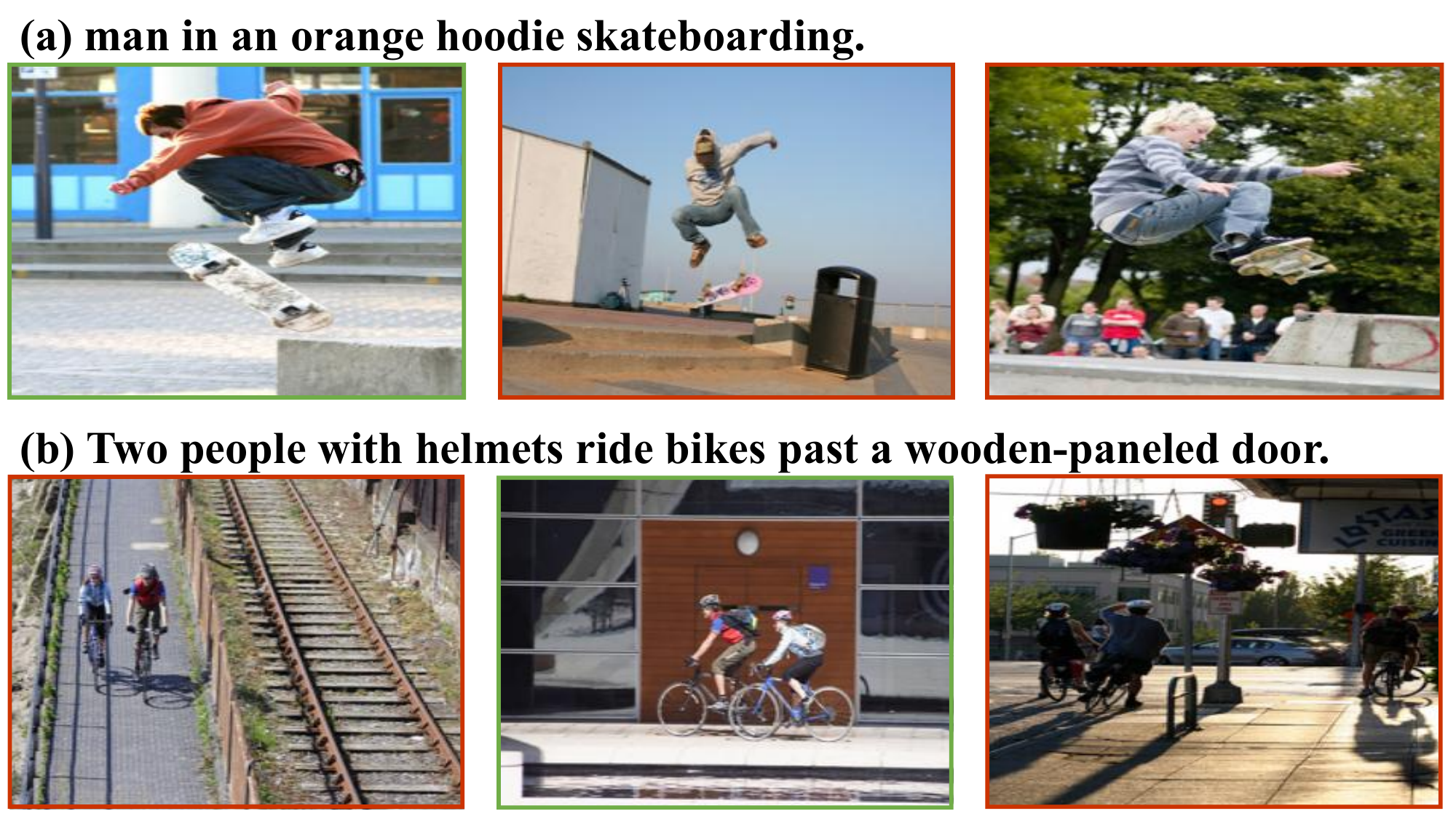}
    \caption{Text-to-Image matching results on Flickr30K.}
    \label{fig:case_2}
\end{figure}

\subsection{Visualization}

To further illustrate effectiveness of TSVC, we present the retrieval process using textual and visual queries, as depicted in Fig.~\ref{fig:case_1} and Fig.~\ref{fig:case_2}. The figures demonstrate how 
TSVC retrieves the top 3 similar images or texts. Even when the ground truth image or text does not rank first, the selected items still capture elements relevant to the query, maintaining consistency in scene or topic. This highlights TSVC’s capability to capture the essential features with subtle comprehension across different modalities.

\section{Conclusion}
In this work, we proposed a Tripartite Learning with Semantic Variation Consistency\,(TSVC) framework to address noisy correspondences for cross-modal data. TSVC leverages information variation between clean image-text pairs to estimate soft correspondence noise labels. By distributing tasks among three models, TSVC effectively resolves the challenge of limited performance improvement due to model prediction homogenization. Extensive experiments on three datasets validate the effectiveness of our method.

\section{Acknowledgments}
This work was supported by the National Natural Science Foundation of China (Grant No. 62076035) and the SMP-Zhipu.AI Large Model Cross-Disciplinary Fund. Special thanks to the State Grid Hebei Electric Power Company (Project No. A2024160) and the China Unicom Beijing Branch Digitalization Department AI Capability Center (Project No. A2024054) for their support.


\bibliography{aaai25.bib}
\end{document}